\documentclass[sn-mathphys-num]{sn-jnl}

\usepackage{adjustbox}
\usepackage{array}
\usepackage{amsmath,amssymb,amsfonts} 
\usepackage{graphicx} 
\usepackage[table,xcdraw]{xcolor} 
\usepackage{booktabs} 
\usepackage{multirow} 
\usepackage{algorithm} 
\usepackage{algorithmic} 
\usepackage{hyperref}

\newcommand{\frameworkname}{{TransNAS-TSAD}}

\begin{document}

\title[\frameworkname{}:  Harnessing Transformers for TSAD]{\frameworkname{}: Harnessing Transformers for Multi-Objective Neural Architecture Search in Time Series Anomaly Detection}

\author*[1]{\fnm{Ijaz} \sur{Ul Haq}}\email{ihaq@uvm.edu}
\author[1]{\fnm{Byung} \sur{Suk Lee}}\email{bslee@uvm.edu}
\author[2]{\fnm{Donna} \sur{M. Rizzo}}\email{drizzo@uvm.edu}
\affil*[1]{\orgdiv{Department of Computer Science}, \orgname{University of Vermont}, \orgaddress{\street{85 Main St}, \city{Burlington}, \postcode{05405}, \state{VT}, \country{USA}}}
\affil*[2]{\orgdiv{Department of Civil and Environmental Engineering}, \orgname{University of Vermont}, \orgaddress{\street{85 Main St}, \city{Burlington}, \postcode{05405}, \state{VT}, \country{USA}}}

\abstract{The surge in real-time data collection across various industries has underscored the need for advanced anomaly detection in both univariate and multivariate time series data. This paper introduces \frameworkname{}, a framework that synergizes the transformer architecture with neural architecture search (NAS), enhanced through NSGA-II algorithm optimization. This approach effectively tackles the complexities of time series data, balancing computational efficiency with detection accuracy. Our evaluation reveals that \frameworkname{} surpasses conventional anomaly detection models due to its tailored architectural adaptability and the efficient exploration of complex search spaces, leading to marked improvements in diverse data scenarios. We also introduce the Efficiency-Accuracy-Complexity Score (EACS) as a new metric for assessing model performance, emphasizing the balance between accuracy and computational resources. \frameworkname{} sets a new benchmark in time series anomaly detection, offering a versatile, efficient solution for complex real-world applications. This research highlights the \frameworkname{} potential across a wide range of industry applications and paves the way for future developments in the field.}

\keywords{Neural Architecture Search, Time Series Anomaly Detection, Transformers, Multi-Objective Optimization}

\maketitle

\footnote{The source code, data, and/or other artifacts have been made available at \url{https://github.com/ejokhan/TransNAS_TSAD.git}.}

\section{Introduction}
\label{intro}

The ubiquity of time series data across various sectors, ranging from finance \cite{bakumenko_2022} and healthcare \cite{samariya_2023} to infrastructure \cite{bhanage_2021} and manufacturing \cite{kammerer_2019}, underscores its pivotal role in modern analytics. These type of data are instrumental in identifying the patterns, dependencies, and anomalies indicative of significant shifts in system behaviors or the emergence of critical issues~\cite{brophy_2023, li_2023}.

Traditionally, statistical methods have been the cornerstone for anomaly detection in time series data, and are renowned for their robust mathematical frameworks~\cite{thudumu_2020, wang_2011, leadbetter_1991}. However, the advent of Big Data, characterized by its significant volume, velocity, and variety, has revealed the limitations of these traditional methods, including high false positive rates and missed detections~\cite{ashabi_2020, al-sai_2019}.

In response, deep learning has introduced a paradigm shift, offering enhanced adaptability and performance in anomaly detection tasks~\cite{ismail_fawaz_2019, ma_2021, haq_2023}. Notably, the transformer architecture, with its self-attention mechanism, has emerged as a groundbreaking development, making significant strides not only in natural language processing but also in time series analysis~\cite{vaswani_2017, tuli_2022, kim_2023}.

Despite the advantages offered by transformer models, their application to time series anomaly detection necessitates a more dynamic and adaptable framework for fine-tuning their structures and parameters for optimal performance~\cite{arslan_2023}. Our model, \frameworkname{}, combines the architectural strengths of transformers (e.g., TranAD) with the optimization strategy of a neural architecture search (NAS) to systematically explore a huge number of architectural configurations. The latter leverages the multi-objective optimization capabilities of the NSGA-II algorithm~\cite{elsken_2019, deb_2002}, enabling an efficient exploration of complex search spaces~\cite{lu_2023, lu_2020}. This
represents a substantial evolution of the TranAD's static architecture, facilitating a balance between performance and computational efficiency.

Adapting anomaly detection models to time series data is inherently complex, especially when addressing multivariate series that present high-dimensional challenges and necessitate sophisticated model architectures~\cite{wei_2018, liu_2018}. The selective attention capabilities of the transformer model offer a compelling solution, provided the architectures are meticulously tailored to the specific research questions and dataset characteristics~\cite{lian_2019}.

The convergence of NAS with multi-objective optimization algorithms like NSGA-II marks a strategic evolution in automated model design. This approach provides a methodical pathway to discovering optimal architectures that consider multiple, often conflicting, objectives, thereby enhancing the adaptability and effectiveness of anomaly detection models~\cite{liu_2021, xue_2023}.

Our research is grounded on three fundamental principles designed to tackle the inherent challenges of time series anomaly detection. First, we emphasize the necessity of developing specialized models tailored to the unique temporal resolution and dimensional characteristics of time series data, ensuring they are finely attuned to the intricacies of temporal patterns. Second, we aim to maximize the utility of transformer architectures, exploiting their advanced capabilities for deep and nuanced interpretation of temporal data, which is critical for identifying subtle anomalies. Lastly, strategic optimization through Neural Architecture Search (NAS), particularly leveraging the NSGA-II algorithm, forms the cornerstone of our methodology. This approach allows us to systematically refine and perfect our models, ensuring they are not only highly effective but also optimized for the specific demands of the datasets they analyze. Together, these principles guide our pursuit of creating models that are both innovative and highly adapted to the complex domain of time series anomaly detection.
Our contributions are delineated as follows:


\begin{itemize}
    \item We introduce a novel approach that applies NAS with NSGA-II optimization to specifically fine-tune transformer architectures for time series anomaly detection, showcasing NAS's adaptability in specialized domains.
    \item Our experimentation identifies key architectural configurations that elevate detection performance, highlighting the importance of architectural fine-tuning in enhancing model efficacy.
    \item We demonstrate substantial performance improvements across diverse datasets, proving our optimized models' superiority over existing methods and illustrating the benefits of multi-objective optimization for anomaly detection.
    \item We strike an optimal balance between computational efficiency and analytical accuracy, presenting models that excel in real-time anomaly detection and are feasible for practical deployment, emphasizing efficiency in model design.
    \item The synergy between transformers' structural strengths and NAS's adaptive optimization creates uniquely effective models for time series anomaly detection, marking a significant advancement by integrating architectural innovation with strategic optimization.
\end{itemize}


\section{Related Work}

The quest to detect and understand anomalies within time series data spans various domains including, but not limited to, finance \cite{bakumenko_2022}, healthcare \cite{samariya_2023}, infrastructure \cite{bhanage_2021}, and manufacturing \cite{kammerer_2019}. This widespread interest underscores the fundamental role of anomaly detection in predictive maintenance, fraud detection, and system health monitoring, among other critical applications \cite{brophy_2023, li_2023}. In finance, for example, anomaly detection can signal fraudulent transactions or market manipulations, while in healthcare, it might indicate abnormal patient conditions that require immediate attention \cite{bakumenko_2022}. 


Initially, the field of time series anomaly detection heavily relied on statistical methods, grounded in robust mathematical frameworks, that have provided deep insights into data patterns and anomalies \cite{leadbetter_1991,wang_2011}. These methods, while effective for simpler datasets, often assume specific data distributions or employ distance-based metrics to identify outliers. However, their application to the high-dimensional, complex nature of multivariate time series data reveals significant limitations. The primary challenge lies not merely in capturing the correlations within the data, but in effectively managing the intricacies associated with multi-correlated variables. These complexities can lead to an increased rate of false positives and missed detections, as traditional methods struggle to adequately interpret the nuanced interactions and dependencies that characterize modern datasets \cite{hodge_austin_2004,patcha_park_2007}.


The emergence of deep learning models, and their ability to learn high-level representations from massive amounts of data, have shown remarkable success in detecting anomalies in time series data \cite{ma_2021}. These models excel in identifying deviations from frequently occurring patterns, significantly reducing false positives and improving detection accuracy \cite{lee_2022, landauer_2023, al-amri_2021}. A deeper exploration of machine learning models, including specific architectures and their successes across various applications, would provide valuable insights \cite{liu_2021}.

Among the deep learning architectures that have gained prominence, the transformer architecture introduced by \cite{vaswani_2017} has revolutionized how models process sequential data. Unlike its predecessors, such as RNNs and LSTMs, the transformer leverages a self-attention mechanism, enabling it to capture long-range dependencies and temporal patterns in data without being hindered by the sequential processing bottleneck \cite{li_2019,tang_matteson_2021,karita_2019,reza_2022,katrompas_2022}. Its application in time series anomaly detection is particularly promising, offering a new perspective on modeling temporal anomalies. However, this area remains relatively unexplored, with significant potential for optimization and adaptation to the unique characteristics of time series data across different domains \cite{thudumu_2020}.

The Neural Architecture Search (NAS) has emerged as a transformative approach in automating the design of neural network architectures \cite{ying_2020,lu_2020,borchert_2022}. By systematically exploring a vast space of architectural configurations, NAS aims to identify models that achieve optimal performance for a given task. This automation is crucial in deep learning, where the design and tuning of models are both resource-intensive and require specialized knowledge \cite{liu_2021,chen_2019,chu_2020}. Recent advancements in NAS have focused on improving its efficiency, reducing the computational resources required for the search process, and making state-of-the-art model architectures accessible for a broader range of applications \cite{lu_2021-3,ying_2020}.

Incorporating multi-objective optimization into NAS, using algorithms like NSGA-II, represents a significant leap forward \cite{deb_2002,chu_2020}. For time series anomaly detection, where models must process large volumes of data accurately and with computational efficiency, the ability to find a nuanced balance between these competing objectives is crucial \cite{wang_2020,lu_2019,ying_2020}. Despite the clear advantages, the application of multi-objective optimization in NAS for anomaly detection in time series is ripe for exploration and offers vast potential for groundbreaking research \cite{liu_2021}.

Our work seeks to harness the strengths of Neural Architecture Search (NAS) and transformer models to directly address the unique challenges associated with temporal data, crafting models that are finely tuned for anomaly detection in time series. This integration aims to develop architectures that excel in both performance and efficiency, leveraging the latest advancements in the field \cite{chitty-venkata_2022}. Given the nascent stage of combining NAS with transformer models for this purpose, our research not only contributes to the existing body of knowledge but also opens avenues for future exploration and innovation in anomaly detection methodologies \cite{wen_2022}. By situating our study within the context of ongoing research efforts and emerging trends, we aim to advance the state of the art in anomaly detection within time series data and contribute meaningfully to the broader discourse in this rapidly evolving field.

\section{Methodology} \label{sec:Methodology}

In this section, we delineate the comprehensive methodology employed in our study, designed to effectively detect and analyze anomalies in multivariate time series data using the \frameworkname{} framework. We commence with a clear definition of the problem, setting the stage for the subsequent methodological steps. This is followed by an in-depth discussion on data refinement techniques, ensuring that the data are appropriately pre-processed for our analysis. Next, we present the intricacies of our transformer architecture, highlighting its design and capabilities. A critical component of our approach is the neural architecture search (NAS), which employs the NSGA-II algorithm for optimizing the transformer model. That description is supplemented by an overview of the evolutionary process that iteratively refines our model architecture. We then transition to discussing advanced anomaly detection techniques, integrating adversarial elements into our model. The section culminates with a description of how we harness the full potential of \frameworkname{} for practical anomaly detection applications, illustrating the real-world implications of our research.

\subsection{Problem Definition}

Our research pioneers the application of a multi-objective approach to anomaly detection in multivariate time series data, utilizing transformer models. The problem is defined as:

Given a sequence of multivariate time series data $S = \{x_1, x_2, \ldots, x_T\}$, where each point $x_t$ is timestamped $t$ and resides in an $m$-dimensional space ($x_t \in \mathbb{R}^m$), our aims are to perform:

\begin{enumerate}
    \item \textbf{Anomaly Detection in Time Series:} Using a training time series $S$ and a test series $S'$ of length $T'$ with similar modalities, we seek to detect anomalies by predicting a binary sequence $Y = \{y_1, y_2, \ldots, y_{T'}\}$, where $y_t$ identifies anomalies at timestamp $t$ in $S'$ (1 indicates an anomaly).
    \item \textbf{Anomaly Component Analysis:} Our goal extends to identifying anomalous components within each data point of the series, generating a detailed prediction sequence $Y = \{y_1, y_2, \ldots, y_{T'}\}$, where each $y_t$ pinpoints the specific anomalous dimensions at timestamp $t$.
\end{enumerate}

\paragraph{Multi-Objective Transformer Model Optimization}

At the heart of our methodology is the use of the NSGA-II based neural architecture search (NAS) for optimizing the transformer architecture, focusing on:

\begin{itemize}
    \item \textbf{Maximizing Anomaly Detection Accuracy:} We aim to enhance the classification performance (F1 score) for precise anomaly detection.
    \item \textbf{Minimizing Architectural Complexity:} Concurrently, reducing the model's size and computational demands for efficiency in practical deployment.
\end{itemize}

This dual-focused approach is a significant advancement in applying transformer models for anomaly detection.

\paragraph{Data Processing and Model Training}

Prior to model training and classification, the time series data undergo a preprocessing phase where normalization is applied to ensure uniformity of scale across all series components. The normalization process is as follows:
\[ x_t \leftarrow \frac{x_t - \min(S)}{\max(S) - \min(S) + \epsilon}, \]
where \( \min(S) \) and \( \max(S) \) denote the component-wise minimum and maximum values observed in the training and testing time series data \( S \), respectively. The term \( \epsilon \) is a small constant introduced to prevent division by zero, ensuring numerical stability.

Subsequently, the time series is transformed into a set of overlapping windows, which serve as the input to our model. A context window of size \( K \) is defined around each data point \( x_t \), resulting in:
\[ W_t = \{x_{t-K+1}, \dots, x_t\}, \]
where boundary cases are handled by replicating the first available data point to fill the window, maintaining a fixed size of \( K \). This approach encapsulates the temporal dependencies inherent in time series data.

\paragraph{Anomaly Scoring and Thresholding}

Our model is designed to assign an anomaly score to each context window \( W_t \), as opposed to direct binary labels. This score is derived from the reconstruction error of the window, where \( O_t \) denotes the reconstructed output corresponding to \( W_t \). A dynamic threshold \( D \) is established based on the distribution of the scores computed on the training data, facilitating the distinction between normal and anomalous windows. The precise mechanism for determining the dynamic threshold \( D \) and how it relates to the detection accuracy are described in subsequent sections.

\subsection{\frameworkname{} Transformer Architecture}

In the \frameworkname{} framework, the transformer architecture is specially adapted for time series anomaly detection. This involves a unique configuration of the standard transformer model to align with the specific characteristics of the time series data.

\paragraph{Data Augmentation Mechanisms Within the Model:} \frameworkname{} integrates specific data augmentation techniques within the model architecture. These techniques, which can be fine-tuned during training, include:

\begin{itemize}
    \item Gaussian Noise Augmentation (Optional): Adds Gaussian noise to the input sequence for variability and robustness.
    \item Time Warping \& Time Masking Augmentation (Optional): Applies time warping or masking, enhancing the model's generalization capabilities.
\end{itemize}

\paragraph{Encoder Operations} The encoder in \frameworkname{} is designed with the following operations:

\begin{itemize}
    \item Linear Embedding (Optional): Transforms the input via linear mapping to a predefined dimensional space.
    \item Layer Normalization: Normalizes the sequence for stability in subsequent operations.
    \item Positional Encoding: Uses either sinusoidal or Fourier positional encoding, depending on the architecture optimized through NAS.

\item Multi-Head Self Attention: A key feature in \frameworkname{} is the multi-head attention mechanism, which is aligned with the number of features in the data, inspired by the TranAD model. It employs the standard scaled-dot product attention mechanism:

\begin{equation}
    \text{Attention}(Q, K, V) = \text{softmax}\left(\frac{QK^T}{\sqrt{d_{\text{model}}}}\right)V.
\end{equation}

The multi-head attention mechanism is formulated as:

\begin{equation}
    \text{MultiHeadAtt}(Q, K, V) = \text{Concat}(H_1, H_2, \ldots, H_h),
\end{equation}
\noindent where $H_i = \text{Attention}(Q_i, K_i, V_i)$.

\item Feedforward Neural Networks (FFNs) Post-Attention: Following the attention mechanism, the sequence is processed through a series of feedforward neural networks (FFNs). Each FFN layer, consisting of a linear transformation followed by an activation function, introduces non-linearities crucial for complex pattern recognition. The configuration of these FFN layers, particularly the number of layers and the number of neurons within each layer, is dynamically determined within a predefined range in our NAS search space. This optimization, conducted using Optuna, is designed to balance anomaly detection performance with computational efficiency, ensuring that the FFN dimensions are optimally tuned for the specific demands of time series anomaly detection.
\end{itemize}

\paragraph{Decoder Operations} The decoder mirrors the encoder's complexity and includes additional transformations that integrate encoder memory, apply multi-head self-attention, and utilize feedforward neural networks, culminating in a final layer for accurate reconstruction. \frameworkname{} can utilize either a dual sequential decoder setup or an iterative approach, depending on the optimized configuration for maximum anomaly detection accuracy.

\begin{figure}[ht]
\centering
\includegraphics[width=0.7\textwidth]{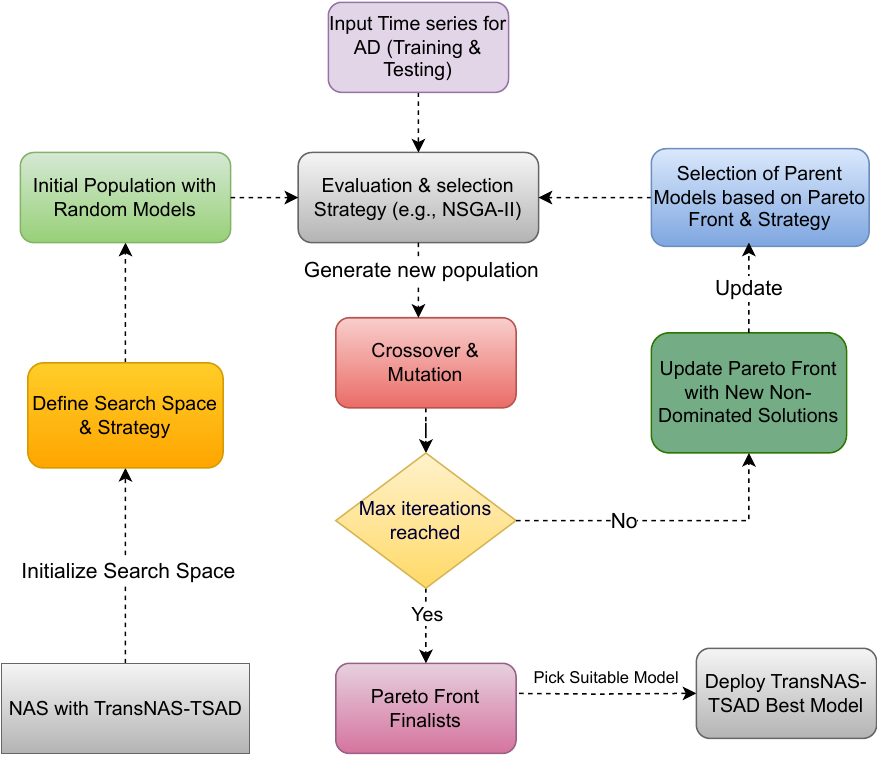}\\
\caption{
Workflow of the \frameworkname{} Process for Time Series Anomaly Detection using Multi-Objective Neural Architecture Search (NAS) with NSGA-II Optimization.}
\label{fig:Framwork}
\end{figure}

\subsection{Neural Architecture Search (NAS) Framework for \frameworkname{}}
\label{sec:nas_framework_transnas_tsad}

The NAS framework shown in Figure \ref{fig:Framwork} is pivotal in achieving our primary objective of effectively detecting anomalies in multivariate time series data. By employing the NSGA-II algorithm, our NAS process systematically navigates the vast architectural landscape of transformer models. This step is essential in identifying an architecture that not only excels in anomaly detection accuracy but also aligns with the computational constraints of practical deployment scenarios. The optimal selection of transformer architectures through NAS directly contributes to enhancing the performance and efficiency of the \frameworkname{} framework, thereby fulfilling our goal of developing a robust and adaptable anomaly detection system.

\subsubsection{Search Space Definition}
The search space within \frameworkname{} and outlined in Table \ref{tab:search_space}  represents a diverse and extensive array of potential model architectures and hyperparameters. This assortment is meticulously tailored toward neural network configurations designed specifically for time-series anomaly detection:

\begin{itemize}
    \item \textit{Architectural Parameters:} These parameters encompass the number and types of layers in the encoder and decoder, optimized attention mechanisms for temporal data analysis, dimensions of feedforward networks, and a variety of positional encoding and normalization methods. Collectively, they enable the model to adapt its architecture for different characteristics of time-series data.
    \item \textit{Training Hyperparameters:} This category includes a broad spectrum of hyperparameters, such as learning rates, batch sizes that are adaptable to varying computational resources, and dropout rates to ensure effective regularization. Additionally, the window size parameter, ranging from 10 to 30, is crucial in defining the length of the input sequence window, which directly influences how the model processes and learns from temporal patterns. Data augmentation techniques like time warping and masking are also part of the search space, enabling the model to simulate and learn from varied anomaly scenarios effectively.
\end{itemize}

The comprehensive nature of the search space in \frameworkname{} allows for the exploration and optimization of a wide range of models, ensuring that the most suitable architecture and hyperparameter settings are identified for effective anomaly detection in different time-series datasets.

\begin{table}[h!]
    
    \centering
    
    \footnotesize
    \begin{adjustbox}{width=\textwidth, center, captionabove={Neural Architecture Search Space for \frameworkname{}}, label=tab:search_space, nofloat=table}
       \begin{tabular}{>{\raggedright\arraybackslash}p{2.5cm} >{\raggedright\arraybackslash}p{3.5cm} p{5cm} p{6.5cm}}
        \toprule
        \textbf{Parameter Type} & \textbf{Parameter Name} & \textbf{Search Space} & \textbf{Description} \\
        \midrule
        \multirow{8}{*}{Training} 
        & Learning Rate & $1 \times 10^{-5}$ to $1 \times 10^{-1}$ (log scale) & Rate at which the model learns. \\
        & Dropout Rate & 0.1 to 0.5 & Regularization to prevent overfitting. \\
        & Batch Size & 16 to 128 (step of 16) & Number of samples per training step. \\
        & Gaussian Noise & $1 \times 10^{-4}$ to $1 \times 10^{-1}$ (log scale) & Noise added for robustness. \\
        & Time Warping & True, False & Augmentation technique for time series. \\
        & Time Masking & True, False & Augmentation technique to mask intervals. \\
        & Window Size & 10 to 30 & Length of the input sequence window for the model. \\
        \midrule
        \multirow{11}{*}{Architectural} 
        & Positional Encoding Type & Sinusoidal, Fourier & Encoding type for sequence position. \\
        & Dimension Feedforward & 8 to 128 (log scale) & Size of the feedforward network. \\
        & Encoder Layers & 1 to 3 & Number of layers in the encoder. \\
        & Decoder Layers & 1 to 3 & Number of layers in the decoder. \\
        & Activation Function & ReLU, Leaky ReLU, Sigmoid, Tanh & Non-linearity after each layer. \\
        & Attention Type & Scaled Dot Product & Type of attention mechanism. \\
        & Number of Attention Heads & Equal to feature dimension & Parallel attention layers. \\
        & Use Linear Embedding & True, False & Option to use a linear embedding layer. \\
        & Layer Normalization & Layer, Batch, Instance & Type of normalization used. \\
        & Self Conditioning & True, False & Conditioning strategy for the model. \\
        & Number of FFN Layers & 1 to 3 & Layers in the feedforward network. \\
        & Phase Type & 1phase, 2phase, Iterative & Model's reconstruction and refinement strategy. \\
        \bottomrule
    \end{tabular}  
\end{adjustbox}
\end{table}

\subsubsection{\frameworkname{}  Evaluation Strategy}
The evolutionary optimization strategy of \frameworkname{}  entails a rigorous evaluation of models generated with varying architectural and training hyperparameters. The multi-objective evaluation focuses on two critical aspects: the F1 score and the number of model parameters. The F1 score serves as a key indicator of the model's accuracy in anomaly detection, balancing precision and recall, while the number of parameters gauges the model's architectural complexity and computational efficiency. The objective is to identify models that not only demonstrate high proficiency in accurately detecting anomalies (as reflected in a high F1 score) but also maintain a streamlined architecture (evidenced by a lower count of parameters). This dual-criteria assessment ensures the selection of models that are both effective in performance and practical in deployment, aligning with the overarching goal of achieving optimal anomaly detection with computational resourcefulness.

\paragraph{Multi-Objective Optimization with NSGA-II}
\frameworkname{} employs the NSGA-II algorithm, a proven leader in balancing the objectives of performance and computational efficiency in multi-objective optimization. Our choice of NSGA-II is substantiated by recent benchmark studies, such as those presented in \cite{lu_2023}, which demonstrate its superiority in identifying optimal solutions across various search spaces. This evidence supports NSGA-II as an effective approach for optimizing the complex and competing demands of our model architecture. The optimization process in \frameworkname{} is governed by the following equations:

\begin{equation}
\mathbf{F}(\mathbf{x}) = [f_1(\mathbf{x}), f_2(\mathbf{x})],
\end{equation}
\noindent where \( f_1(\mathbf{x}) \) is the F1 score and \( f_2(\mathbf{x}) \) is a computational resource metric.

Non-dominated sorting and the calculation of crowding distance are integral parts of NSGA-II:

\begin{equation}
\text{NonDomSort}(\mathbf{X}) = \bigcup_{i=1}^{k} F_i
\end{equation}

\begin{equation}
d(\mathbf{x}) = \sum_{i=1}^{k} \left( f_i(\mathbf{x}^{+}) - f_i(\mathbf{x}^{-}) \right)
\end{equation}

\paragraph{Pareto Front Exploration and utilization}
The Pareto front represents the set of non-dominated solutions, providing an optimal trade-off between conflicting objectives:

\begin{equation}
\mathcal{P} = \{ \mathbf{x} \in \mathbf{X} \mid \nexists \mathbf{y} \in \mathbf{X} : \mathbf{F}(\mathbf{y}) \prec \mathbf{F}(\mathbf{x}) \},
\end{equation}
\noindent where \( \prec \) indicates that \( \mathbf{y} \) dominates \( \mathbf{x} \).

The Pareto front derived from the NAS process provides a practical framework for model selection:

\begin{itemize}
    \item \textit{Interpretation and Analysis:} The front is analyzed to discern architectural trade-offs, allowing stakeholders to identify models with the desired balance 
    between accuracy and efficiency.
    \item \textit{Informed Selection:} Decision-makers can select models that align with specific performance expectations and operational constraints by examining the Pareto front.
    \item \textit{Resource Allocation:} Models on the Pareto front inform resource allocation, directing investments toward architectures with the most favorable trade-offs.
\end{itemize}

Algorithm \ref{alg:\frameworkname{}} provides a detailed procedure that encapsulates our approach within the Optuna framework using NSGA-II \cite{deb_2002} for the multi-objective optimization. This algorithm highlights, from initialization to final selection,  the integration of the evolutionary search process with transformer-specific architecture and hyperparameter tuning.

\begin{algorithm}[!ht]
\caption{Multi-Objective NAS for Time Series Anomaly Detection with \frameworkname{}}\label{alg:\frameworkname{}}
\begin{algorithmic}[1]
\STATE \textbf{Input:} Training dataset, validation dataset, predefined search space for transformer architectures and hyperparameters
\STATE \textbf{Output:} Optimized transformer architecture and hyperparameters for time series anomaly detection
\STATE \textbf{Initialization:} Set up the Optuna framework with the NSGA-II sampler. Initialize the search with a population of transformer architectures from the predefined search space. Define evolutionary process parameters: population size, number of generations, mutation and crossover rates.
\STATE Specify evaluation metrics: F1 score, computational resource utilization (model size, training time), precision, recall, and inference time.
\FOR{each study in Optuna}
    \FOR{each trial in the study}
        \STATE Define the transformer model architectural parameters and training hyperparameters within the search space constraints.
        \STATE Construct and train the transformer model on the Training dataset.
        \STATE Evaluate the model on the Validation dataset using the specified metrics.
        \STATE Update the study with the trial's objectives, reflecting the model's performance and computational efficiency.
        \IF{early stopping criteria are met based on validation performance or predefined resource constraints}
            \STATE Terminate the trial prematurely.
        \ENDIF
    \ENDFOR
    \STATE \textbf{Selection:} Use Optuna's NSGA-II sampler to select architectures for the next generation, focusing on the Pareto-optimal solutions.
    \STATE \textbf{Crossover and Mutation:} Apply Optuna's genetic operators to evolve the study's population, exploring new architectural variants.
\ENDFOR
\STATE \textbf{Final Selection:} At the end of the NAS process, use Optuna to extract the best transformer architecture and hyperparameters from the Pareto front, considering both performance metrics and computational efficiency.
\end{algorithmic}
\end{algorithm}

\subsubsection{Evolutionary Process in \frameworkname{}}
\label{sec:evolutionary_process_transnas_tsad}
The evolutionary process within \frameworkname{} is critical in achieving our overarching goal of refining anomaly detection capabilities. Through iterative refinement and selection of model architectures, this process ensures that our models are not only accurate in identifying anomalies but also evolve to become more efficient and adaptable to various data characteristics. This continuous improvement and adaptation are key to achieving high-performance anomaly detection, which stands at the core of our research objectives.

\begin{itemize}
    \item \textit{Generation Loop:} New models are generated, evaluated, and passed on to each generation, with superior architectures refining the Pareto front.
    \item \textit{Early Stopping:} The process incorporates early stopping for models that do not show promise, thus conserving computational resources.
    \item \textit{Final Architecture Selection:} The optimal architecture is selected from the Pareto front, ensuring an effective balance between the objectives.
\end{itemize}

\subsection{Offline Optimization of Transformer Architecture for Anomaly Detection}
\label{sec:offline_optimization_transformer}

Central to \frameworkname{} is the offline, strategic optimization of the transformer architecture, a process critical for achieving an effective balance between anomaly detection accuracy and computational efficiency. This optimization, guided by neural architecture search (NAS) using the NSGA-II algorithm, involves:

\begin{enumerate}
    \item \textit{Comprehensive Trial-Based Exploration:} In line with Optuna's best practices, the process involves conducting over 100 trials to explore a wide array of transformer architectures, ensuring a thorough examination of the solution space.
    \item \textit{Achieving Pareto Front Efficiency:} The key focus of NAS is to identify and refine a collection of architectures that constitute the Pareto front, representing a spectrum of high-performance, resource-intensive models to more balanced and resource-efficient alternatives.
    \item \textit{Iterative Evolutionary Process:} Through evolutionary algorithms, these architectures undergo continuous adaptation and improvement, exploring innovative configurations and enhancing performance and efficiency with each generation.
\end{enumerate}

\paragraph{Tailoring Architectures to Specific Deployment Needs}
\label{subsec:tailoring_architectures_deployment}

Post the offline NAS process, we are equipped with a diverse array of transformer architectures, each optimized for different operational contexts:

\begin{itemize}
    \item \textit{For Resource-Rich Environments:} High F1 score models, though computationally demanding, are suitable for scenarios where resource availability is not a constraint.
    \item \textit{In Resource-Limited Settings:} The NAS process also yields architectures that are either balanced in terms of performance and resource usage or are specifically optimized for low-resource environments.
\end{itemize}

This flexibility allows for the deployment of \frameworkname{} in various settings, ensuring efficient and effective anomaly detection tailored to the constraints and requirements of different deployment environments.

\subsection{Advanced Anomaly Detection with Adversarial Elements in \frameworkname{}}
\label{sec:advanced_anomaly_detection}



\frameworkname{} represents a breakthrough in the field of time series anomaly detection by strategically enhancing adversarial learning paradigms. While it draws foundational inspiration from the TranAD model's two-phase approach, our framework introduces: a neural architecture search (NAS) strategy that effectively incorporates a third, iterative self-adversarial phase. This tripartite approach enables dynamic selection among the conventional two-phase mechanism and our advanced iterative phase, optimizing detection capabilities for each specific trial.

The inception of this third phase signifies a substantial evolution in anomaly detection techniques. By embedding adversarial elements within all three phases, \frameworkname{} not only retains the strengths of the traditional encoder-decoder models but also introduces a level of adaptability and precision previously unattainable. This method excels in identifying subtle and complex anomalies, leveraging the iterative adversarial training to refine detection with each iteration.


\subsubsection{Three-Phase Adversarial Approach}

Drawing from adversarial learning paradigms, the three-phase approach in \frameworkname{} seeks to enhance the model's sensitivity to anomalies through competitive reconstruction stages.

\paragraph{Phase 1 - Preliminary Input Reconstruction}
This phase, acting as the foundation, aims for a preliminary reconstruction of the input time-series window, and yields a focus score. The latter is defined by the deviation of the reconstructed output from the actual input:
\[
L_{\text{focus}} = \Vert O_{\text{initial}} - W \Vert^2,
\]
\noindent where \( O_{\text{initial}} \) is the output from the first phase and \( W \) is the input time-series window. The resultant ``focus score'', derived from the deviations in this initial reconstruction, serves as an attention modulator for the subsequent phase.

\paragraph{Phase 2 - Adversarial, Focus-Driven Reconstruction}
Incorporating adversarial elements, this phase utilizes the focus scores from phase 1. In the adversarial phase, the second decoder aims to maximize the difference between its output and the input. Simultaneously, the first decoder aims to minimize this difference:
\[
L_{\text{adv1}} = \Vert O_{\text{adv1}} - W \Vert^2
\]
\[
L_{\text{adv2}} = -\Vert O_{\text{adv2}} - W \Vert^2,
\]
\noindent where \( O_{\text{adv1}} \) and \( O_{\text{adv2}} \) are the outputs from the first and second decoders, respectively, during the adversarial phase.

\paragraph{Phase 3 -Iterative Self-Adversarial Approach}

Going beyond the structured two-phase reconstruction inspired by the TranAD work, \frameworkname{} introduces a dynamic iterative approach. This approach, embedded with self-adversarial mechanisms, continually refines its reconstructions.

\subparagraph{Iterative Refinement:}

Starting with an initial reconstruction, the deviation between the current output and the input provides feedback for the next iteration:

\[
L_{\text{iteration}} = \left\| O_{\text{current}} - W \right\|_2
\]

The self-adversarial mechanism can be represented as:

\[
L_{\text{self-adv}} = (L_{\text{iteration, prev}} - L_{\text{iteration, current}})^2,
\]
\noindent where \( L_{\text{iteration, prev}} \) is the loss from the previous iteration and \( L_{\text{iteration, current}} \) is the loss from the current iteration. The iterative refinement continues until the change in the loss between consecutive iterations falls below a predetermined threshold, signifying convergence. Mathematically, the convergence criterion can be defined as:

\[
\Delta L = \left| L_{\text{iteration, current}} - L_{\text{iteration, prev}} \right|.
\]

If \( \Delta L < \epsilon \), where \( \epsilon \) is a small positive value (e.g., \( 10^{-5} \)), the iteration stops, suggesting that further refinement may not yield significant improvements.
After the iterative process converges, the best anomaly score is determined. The best score is derived from the iteration with the smallest reconstruction loss, ensuring that the selected representation most closely matches the input time-series data.

\paragraph{Adaptable and Robust Anomaly Detection}
What sets \frameworkname{} apart is its adaptability. The model is not confined to a fixed number of reconstruction phases. Depending on the intricacy of the dataset, it can dynamically adjust, ensuring that even the most subtle anomalies are not overlooked. Moreover, the incorporation of adversarial elements, both in the two-phase and iterative approaches, ensures the model continually challenges itself, leading to more refined and accurate reconstructions. 
\frameworkname{} represents a significant advancement in the realm of time-series anomaly detection. By amalgamating auto-regressive inference, adversarial mechanisms, attention modulation, and adaptive iterations, it offers a comprehensive solution, adept at detecting both overt and nuanced anomalies across varied time-series datasets.

\subsection{Harnessing the Power of \frameworkname{} for Anomaly Detection}
Building upon the foundational architecture and adversarial mechanisms of \frameworkname{}, we now delve into the practical realm of anomaly detection. This section elucidates the methodologies employed to infer anomalies from time-series data, leveraging the trained transformer model and various augmentation strategies.

\paragraph{The Anomaly Scoring Mechanism}

Every incoming data point, denoted as \( \hat{W} \), is assessed for anomalies by computing a score that quantitatively reflects its deviation from expected patterns. Here, \( \hat{W} \) represents the new window of time series data under consideration. The anomaly score, \( s \), is calculated as:
\begin{equation}
s = \frac{1}{2} \| R_1 - \hat{W} \|_2^2 + \frac{1}{2} \| \hat{R}_2 - \hat{W} \|_2^2,
\end{equation}
where \( R_1 \) and \( \hat{R}_2 \) are the reconstructed outputs from the transformer model, corresponding to different stages of the inference process.

\paragraph{Dual Inference Pathways}
\frameworkname{} boasts of two distinct inference methodologies:
\begin{itemize}
    \item \textit{Two-Phase Approach:} A structured two-step inference process yielding reconstructions $(R_1, \hat{R_2})$.
    \item \textit{Iterative Refinement:} For scenarios demanding intricate attention, we engage in iterative inference, perfecting the anomaly scores with each iteration.
\end{itemize}

\begin{figure}[ht]
\centering
\includegraphics[width=0.7\textwidth]{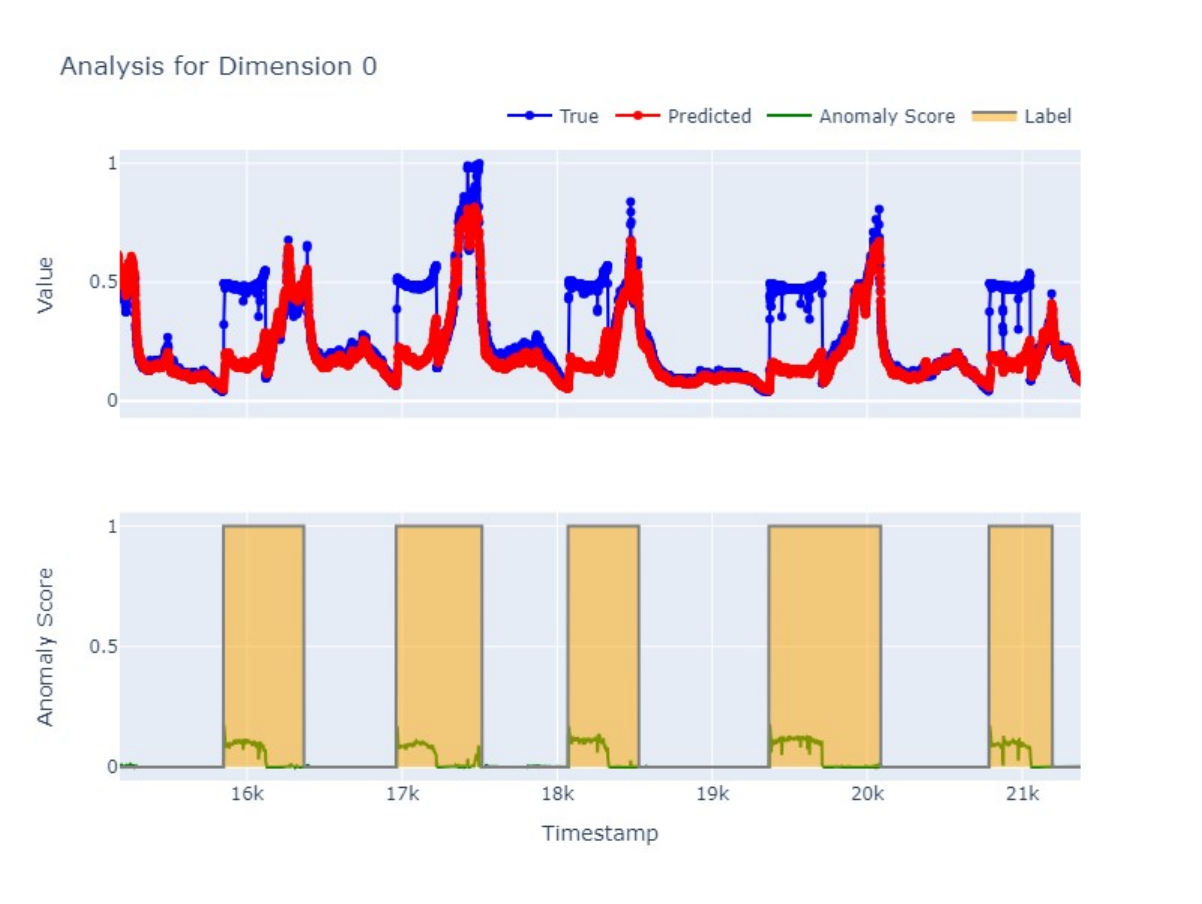}\\
\vspace*{-1em}
\caption{Anomaly detection in \frameworkname{} on dimension 0 of the SMD test dataset, depicting true versus predicted values and identified anomalies}
\label{fig:AnomalyDetection}
\end{figure}

Figure~\ref{fig:AnomalyDetection} illustrates the anomaly detection process on dimension 0 of the Server Machine dataset (SMD) test dataset using \frameworkname{}. The upper plot displays the true data points (in blue) and the predicted values (in red), while the lower plot indicates the computed anomaly scores and their corresponding labels, offering a visual interpretation of the model's detection capabilities in identifying anomalies.

\paragraph{Dynamic Anomaly Thresholding: The Sentinel}
Central to our thresholding strategy is the peaks over threshold (POT) method. At its core, POT establishes a threshold where observations surpassing this mark are deemed anomalous, allowing for effective discernment of extreme data points.

Evolutionary in nature, our modified POT (mPOT) approach ensures adaptability to the ever-changing landscape of time-series data:
\begin{equation}
mPOT(x) = POT(x) + \alpha \times \text{recent\_deviation}(x),
\end{equation}
\noindent where $\alpha$ is a weight factor. The function \text{recent\_deviation} calculates the deviation of the latest data points from their median value. Anomalies are pinpointed whenever any dimension's score, $s_i$, exceeds this dynamic mPOT threshold.

\subsubsection{Augmentative Strategies for Enhanced Precision}
\frameworkname{} employs various augmentative strategies to bolster the precision of the POT method, ensuring that the anomaly detection mechanism remains sensitive, adaptable, and robust. These strategies aim to enhance the detection capabilities by refining the anomaly scores and the thresholds against which they are evaluated.

\paragraph{Moving Average Thresholding (MAT)}
MAT is a dynamic thresholding technique that complements the POT method. Instead of using a static threshold, MAT calculates a moving average of recent anomaly scores to adaptively set the threshold. This dynamic adjustment ensures that the threshold is responsive to emerging data trends and patterns, enhancing its relevance and accuracy. The moving average threshold at time \(t\) is given by:
\begin{equation}
MAT(t) = \frac{1}{N} \sum_{i=t-N}^{t} s_i,
\end{equation}
\noindent where $(s_i$ is the anomaly score at time $i$ and $N$ represents the window size for the moving average.

\paragraph{Rolling Statistics for Nuanced Detection}
Rolling statistics, particularly the rolling mean and standard deviation, provide additional context to the POT method by capturing temporal dependencies and trends in the data. These statistics are instrumental in uncovering subtle anomalies that might otherwise be overlooked. For each data point at time $t$, the rolling mean $\mu(t)$ and rolling standard deviation $\sigma(t)$ are computed as:
\begin{equation}
\mu(t) = \frac{1}{W} \sum_{i=t-W}^{t} x_i \quad \text{and} \quad \sigma(t) = \sqrt{\frac{1}{W} \sum_{i=t-W}^{t} (x_i - \mu(t))^2},
\end{equation}
\noindent where $x_i$ represents the data point at time $i$ and $W$ is the rolling window size. These rolling statistics are incorporated into the feature set, enhancing the model’s ability to discern intricate data variations and improving the relevance of the anomaly scores generated for the POT method.


\section{Experimental Setup for \frameworkname{}}

We evaluate the performance of \frameworkname{} against a suite of established benchmark models, including TranAD \cite{tuli_2022}, LSTM-NDT \cite{hundman_2018}, DAGMM \cite{zong_2018}, OmniAnomaly \cite{su_2019}, MSCRED \cite{zhang_2019}, MAD-GAN \cite{li_2019-2}, USAD \cite{audibert_2020}, CAE-M \cite{zhang_2021}), and GDN \cite{deng_hooi_2021}. To ensure a fair comparison, we utilize the hyperparameter configurations as specified in the original publications of these models, relying on their publicly available implementations.

The experimental infrastructure comprises a Google Colab Pro environment, leveraging a NVIDIA Tesla T4 GPU (16GB memory) and an Intel Xeon E5-2670 v3 CPU (8 cores, 51GB memory). The NSGA-II based NAS algorithm and data management operations are implemented in Python, utilizing robust libraries such as PyTorch for deep learning, Optuna for hyperparameter optimization, and pandas for data handling.



A pivotal element of our methodology is the neural architecture search (NAS), which automates the architectural design by navigating a comprehensive search space of architectural and training hyperparameters, as shown in Table \ref{tab:search_space}. This search space is crafted to enable the NAS algorithm to adapt the model architecture to the unique characteristics of each dataset. Consistent with the practices in OmniAnomaly and TranAD, we apply an enhanced version of the peaks over threshold (POT) method \cite{leadbetter_1991} across all datasets with a uniform coefficient of $10^{-4}$. This enhancement incorporates our previously described augmentative strategies, further fine-tuning the low quantile parameter for each specific dataset to align with established benchmarks and ensure equitable comparative analysis.

\subsection{Datasets}
\begin{table}[h!]
\centering
\footnotesize
\begin{adjustbox}{width=\textwidth, center, captionabove={Dataset Statistics}, label=tab:dataset_statistics, nofloat=table}
\begin{tabular}{@{}lrrrrr@{}}
\toprule
Dataset & \multicolumn{1}{l}{Train Size} & \multicolumn{1}{l}{Test Size} & \multicolumn{1}{l}{Dimensions} & \multicolumn{1}{l}{Sequences} & \multicolumn{1}{l}{Anomalies (\%)} \\ \midrule
NAB     & 4,033                          & 4,033                         & 1                             & 6                             & 0.92                              \\
UCR     & 1,600                          & 5,900                         & 1                             & 4                             & 1.88                              \\
MBA     & 100,000                        & 100,000                       & 2                             & 8                             & 0.14                              \\
SMAP    & 135,183                        & 427,617                       & 25                            & 55                            & 13.13                             \\
MSL     & 58,317                         & 73,729                        & 55                            & 3                             & 10.72                             \\
SWaT    & 496,800                        & 449,919                       & 51                            & 1                             & 11.98                             \\
WADI    & 1,048,571                      & 172,801                       & 123                           & 1                             & 5.99                              \\
SMD     & 708,405                        & 708,420                       & 38                            & 4                             & 4.16                              
                            \\ \bottomrule
\end{tabular}
\end{adjustbox}
\end{table}

Our study employs suite of datasets selected to challenge and validate the robustness of the time series anomaly detection methods across a variety of real-world scenarios.  These datasets, renowned for their complexity and diversity, allow for an effective benchmarking of our \frameworkname{} approach against existing methodologies. See Table \ref{tab:dataset_statistics} for summary statistics.
\begin{itemize}
  
    \item \textbf{Numenta Anomaly Benchmark (NAB) Dataset}: This dataset encompasses a wide range of real-world data, including temperature sensor readings, cloud machine CPU utilization, service request latencies, and taxi demand data in New York City. It's important to note some anomalies in labeling within this dataset are excluded from analyses, particularly in the NYC-taxi traces  \cite{ahmad_2017,nakamura_2020}.
  
    \item \textbf{HexagonML (UCR) Dataset}: Featured in the KDD 2021 cup, this dataset comprises a diverse collection of univariate time series. In our research, we selectively use natural data representations derived from real-world sources, focusing specifically on the InternalBleeding and ECG datasets.  Consequently, we did not include any synthetic sequences that are also part of the HexagonML (UCR) dataset \cite{dau_2018}.

    \item \textbf{MIT-BIH Supraventricular Arrhythmia Database (MBA) Dataset}: This dataset includes electrocardiogram recordings from four patients, featuring anomalies such as supraventricular contractions or premature heartbeats. It's a well-recognized dataset in data management studies \cite{moody_mark_2001,boniol_2020}.  

    \item \textbf {Soil Moisture Active Passive (SMAP) Dataset}: This NASA-provided dataset includes global measurements of soil moisture in the top 5 cm of Earth's soil surface, collected approximately every three days by the SMAP satellite. It is designed to enhance our understanding of water, carbon, and energy cycles\cite{hundman_2018}. 


  \item \textbf{Mars Science Laboratory (MSL) Dataset}: Similar to the SMAP dataset, the MSL dataset includes sensor and actuator data from the Mars rover.\cite{hundman_2018} Due to the presence of many trivial sequences, only specific non-trivial sequences are typically analyzed such as (A4, C2 and T1) pointed by \cite{nakamura_2020,tuli_2022}.

  \item \textbf{Secure Water Treatment (SWaT) Dataset}: This dataset is derived from a real-world water treatment plant, including data from 7 days of normal operations and 4 days under abnormal conditions. It features readings from various sensors and actuators \cite{mathur_tippenhauer_2016}. 
  
  \item \textbf{Water Distribution (WADI) Dataset}: An expansion of the SWaT dataset, WADI includes a larger array of sensors and actuators. The dataset spans a longer period, with 14 days of normal operation and 2 days under attack scenarios \cite{ahmed_2017}.
  
  \item \textbf{Server Machine Dataset (SMD)}: Covering five weeks of data, this dataset includes resource utilization traces from 28 machines in a compute cluster. Only specific non-trivial sequences are used for analysis \cite{su_2019}. 
\end{itemize}


\subsection{Evaluation Criteria}
\label{sec:evaluation_criteria}
We employ key metrics aligned with the objectives of the TransNAS-TSAD framework. These metrics are selected to optimize anomaly detection effectiveness, considering practical deployment aspects.
The F1 score, defined as the harmonic mean of precision and recall, is central to our evaluation strategy. It is a performance measure that balances the trade-off between false positives and false negatives to help address the challenge of imbalanced datasets common in anomaly detection.

\subsubsection{Introducing the Efficiency-Accuracy-Complexity Score (EACS)}

Model complexity is frequently measured by parameter count, a metric that, while useful,doesn’t fully capture the capabilities of transformer-based models like TransNASTSAD. Despite their high parameter count, these models are highly effective at capturing complex data patterns, a strength that justifies their complexity. Their advanced processing capabilities, facilitated by parallelization, make them adept at
handling scenarios that demand high accuracy, even when computational resources are abundant. Therefore, when we report the parameter count of TransNAS-TSAD,
it is within the broader context of its sophisticated functionality. Additionally, our evaluation emphasizes not just parameter count but also the F1 score and Efficiency- Accuracy-Complexity Score (EACS), ensuring our assessment considers both the model’s anomaly detection effectiveness and its deployability. While other metrics like ROC/AUC are valuable, our focused approach aligns with the specific objectives of
TransNAS-TSAD, aiming for a balanced evaluation of its performance.

To conduct a fair and comparative assessment across models, we have calculated the number of parameters, training time, and F1 score for each benchmark model as well as for the best model instance obtained from TransNAS-TSAD for each dataset. The maximum values for F1 score, training time, and parameter count—represented as \( \text{F1}_{\text{max}}, \text{T}_{\text{max}}, \text{P}_{\text{max}} \)—are the highest values observed among all models compared for each specific dataset. This normalization ensures a fair comparison across models. The EACS is defined as:

{\footnotesize
\centering
\begin{equation}
    EACS = w_{a} \times \left( \frac{\text{F1}}{\text{F1}_{\text{max}}} \right) + 
           w_{t} \times \left(1 - \frac{\text{T}_{\text{train}}}{\text{T}_{\text{max}}} \right) + 
           w_{p} \times \left(1 - \frac{\text{P}_{\text{count}}}{\text{P}_{\text{max}}} \right)
\end{equation}
}

where:
- \( w_{a} \) is the weight for the accuracy component (F1 score), set to 0.4,
- \( w_{t} \) is the weight for the training time efficiency component, set to 0.4,
- \( w_{p} \) is the weight for the model complexity (parameter count) component, set to 0.2.

These weights were chosen to reflect the importance of accuracy and training efficiency in practical deployments, where they are often prioritized over the complexity of the model. The F1 score (\( \text{F1} \)) represents the accuracy of the model, \( \text{T}_{\text{train}} \) denotes the training time, and \( \text{P}_{\text{count}} \) indicates the parameter count of the model.


\section{Results}
\label{sec:results_analysis}
\paragraph{Evaluating \frameworkname{}: Precision, Recall, and F1 Score Analysis}

The evaluation of anomaly detection models, as shown in Table \ref{Table:3}, measures \frameworkname{}'s performance across diverse datasets including NAB, UCR, MBA, SMAP, MSL, SWaT, WADI, and SMD, using precision, recall, and F1 scores as benchmarks. \frameworkname{} demonstrated high F1 scores in datasets such as NAB, UCR, MBA, and SMAP, benefiting from its advanced data processing techniques that adeptly handle complex patterns in both univariate and multivariate time series. This distinguishes it from models like OmniAnomaly and MSCRED. For the MSL dataset, the TranAD model slightly outperforms \frameworkname{} (0.9494 vs 0.9482), likely due to TranAD’s parameters being finely tuned to MSL's unique characteristics. Despite conducting over 100 trials per dataset, this suggests that even more exhaustive optimization could potentially unlock further improvements for \frameworkname{}. Its robust performance in the SWaT dataset, with an F1 score of 0.8314, underscores its versatility in various industrial contexts. In the cases of WADI and SMD, where \frameworkname{} achieves F1 scores of 0.8400 and 0.9986 respectively, the significant improvements—such as a 40\% increase over baseline in WADI—are attributable to its comprehensive search space and optimization strategies, enabling effective model tuning. This evaluation underscores the potential benefits of extending our optimization framework to achieve even greater model refinement.

The variable performance of models like LSTM-NDT and DAGMM across different datasets points to their methodological strengths and constraints. For example, the LSTM-NDT's reduced efficacy likely stems from the method's nonparametric thresholding and resulting sensitivity to large differences in anomaly patterns; and the DAGMM struggles with longer sequences, in part, due to its reliance on a singular GRU model, which limits its capacity to accurately map extended temporal dynamics. Moreover, \frameworkname{} demonstrates enhanced performance over innovative yet less adaptable models such as MAD-GAN and USAD, and while CAE-M and GDN show potential in specific scenarios, they generally fall short of matching \frameworkname{}’s applicability and performance across the datasets reviewed, highlighting the value of integrating  adversarial training and iterative optimization into the \frameworkname{}'s framework.

\begin{table}[h!]
\centering
\footnotesize
\begin{adjustbox}{width=\textwidth, center, captionabove={Benchmarking \frameworkname{} against Baseline Models: A summary of precision (P), recall (R), and F1 scores showcases the advanced anomaly detection capabilities of \frameworkname{}, derived via NSGA-II-based NAS, across diverse datasets.}, label=tab:Benchmarking, nofloat=table}
\begin{tabular}{lccc|ccc|ccc|ccc}
\toprule
\multicolumn{1}{c}{\textbf{Method}} & \multicolumn{3}{c}{\textbf{NAB}} & \multicolumn{3}{c}{\textbf{UCR}} & \multicolumn{3}{c}{\textbf{MBA}} & \multicolumn{3}{c}{\textbf{SMAP}} \\
 & P & R & F1 & P & R & F1 & P & R & F1 & P & R & F1 \\
\midrule
LSTM-NDT & 0.6400   & 0.6667  & 0.6531 & 0.5231 & 0.8294 &  0.5231 & 0.9207 & 0.9718 &  0.9456 & 0.8523 & 0.7326 & 0.7879 \\
DAGMM & 0.7622 & 0.7292 &  0.7453 & 0.5337 & 0.9718 & 0.5337 & 0.9475 & 0.9900   & 0.9683 & 0.8069 & 0.9891 & 0.8888  \\
OmniAnomaly & 0.8421 & 0.6667   & 0.7442 & 0.8346 & 0.9999 & 0.8346 & 0.8561 & 1.000  & 0.9225 & 0.813  & 0.9419 & 0.8728  \\
MSCRED & 0.8522 & 0.6700   & 0.7502 & 0.5441 & 0.9718  & 0.5441 & 0.9272 & 1.0000 & 0.9623 & 0.8175 & 0.9216 & 0.8664  \\
MAD-GAN & 0.8666 & 0.7012 & 0.7752 & 0.8538 & 0.9891  & 0.8538 & 0.9396 & 1.0000  & 0.9689 & 0.8157 & 0.9216 & 0.8654  \\
USAD & 0.8421 & 0.6667   & 0.7442 & 0.8952 & 1.0000 & 0.8952 & 0.8953 & 0.9989  & 0.9443 & 0.748  & 0.9627 & 0.8419 \\
CAE-M  & 0.7918 & 0.8019  & 0.7968 & 0.6981 & 1.0000 & 0.6981 & 0.8442 & 0.9997  & 0.9154 & 0.8193 & 0.9567 & 0.8827 \\
GDN & 0.8129 & 0.7872  & 0.7998 & 0.6894 & 0.9988  & 0.6894 & 0.8832 & 0.9892 & 0.9332 & 0.748  & 0.9891 & 0.8518 \\
TranAD & 0.8889 & 0.9892  & 0.9364 & 0.9407 & 1.0000  & 0.9407 & 0.9569 & 1.0000 & 0.978  & 0.8043 & 0.9999 & 0.8915  \\

\frameworkname{} & 0.8888 & 1.0000 &  \textbf{0.9411} & 0.9823 & 1.0000 & \textbf{0.9910} & 0.9726 & 1.0000  & \textbf{0.9861} & 0.9066 & 1.0000 & \textbf{0.9510} \\
\midrule
\multicolumn{1}{c}{\textbf{Method}} & \multicolumn{3}{c}{\textbf{MSL}} & \multicolumn{3}{c}{\textbf{SWaT}} & \multicolumn{3}{c}{\textbf{WADI}} & \multicolumn{3}{c}{\textbf{SMD}} \\
 & P & R & F1 & P & R & F1 & P & R & F1 & P & R & F1 \\
\midrule
LSTM-NDT & 0.6288 & 1.0000 & 0.7721 & 0.7778 & 0.5109  & 0.6167 & 0.0138 & 0.7823 & 0.0271 & 0.9736 & 0.844   & 0.9042  \\
DAGMM & 0.7363 & 1.0000  & 0.8482 & 0.9933 & 0.6879 & 0.8128 & 0.076  & 0.9981 & 0.1412 & 0.9103 & 0.9914 & 0.9491  \\
OmniAnomaly & 0.7848 & 0.9924 & 0.8765 & 0.9782 & 0.6957 & 0.8131 & 0.3158 & 0.6541 & 0.426  & 0.8881 & 0.9985 & 0.9401\\
MSCRED & 0.8912 & 0.9862 & 0.9363 & 0.9992 & 0.677 & 0.8072 & 0.2513 & 0.7319 & 0.3741 & 0.7276 & 0.9974 & 0.8414\\
MAD-GAN & 0.8516 & 0.993  & 0.9169 & 0.9593 & 0.6957 & 0.8065 & 0.2233 & 0.9124 &  0.3588 & 0.9991 & 0.844  & 0.915  \\
USAD & 0.7949 & 0.9912  & 0.8822 & 0.9977 & 0.6879 & 0.8143 & 0.1873 & 0.8296 & 0.3056 & 0.906  & 0.9974 & 0.9495 \\

CAE-M & 0.7751 & 1.0000  & 0.8733 & 0.9697 & 0.6957  & 0.8101 & 0.2782 & 0.7918 & 0.4117 & 0.9082 & 0.9671 & 0.9367 \\
GDN & 0.9308 & 0.9892 & \textbf{ 0.9591} & 0.9697 & 0.6957 & 0.8101 & 0.2912 & 0.7931 & 0.426  & 0.717  & 0.9974 & 0.8342 \\
TranAD & 0.9038 & 0.9999  & 0.9494 & 0.976  & 0.6997  & 0.8151 & 0.3529 & 0.8296  & 0.4951 & 0.9262 & 0.9974 & 0.9605 \\

\frameworkname{} & 0.9567 & 1.000 & 0.94.82 & 0.9415 & 0.7624 & \textbf{0.8314} & 0.8508  & 0.8295 & \textbf{0.8400} & 0.9985 & 0.9988 & \textbf{0.9986} \\
\bottomrule
\end{tabular}
\end{adjustbox}
\end{table}

\paragraph{In-Depth Analysis of Training Efficiency and EACS in Anomaly Detection Models}
\label{sec:results_analysis-EACS}

Table \ref{Tab: EACS} presents a comprehensive analysis that compares the Efficiency-Accuracy-Complexity Score (EACS) among various anomaly detection models, including \frameworkname{}. It highlights how training time and model size significantly influence the application of anomaly detection models in operational environments that demand rapid deployment and updates, a factor underscored by several studies \cite{johnson_2018,nokhwal_2023,coquelin_2022}. In this analysis, \frameworkname{} emerges as a standout performer, achieving high EACS values across multiple datasets, which indicates its effective balance of model performance and operational efficiency. Specifically, \frameworkname{} delivers high F1 scores of 94.11\% and 99.10\% on the NAB and UCR datasets, respectively, while maintaining brief training periods of 2.70 and 2.24 seconds. This translates into EACS values of 0.9742 and 0.9922, showcasing \frameworkname{}'s rapid adaptability and ability to learn from complex data. The trend (i.e., high accuracies combined with short training times) is consistent across other datasets, including MBA and SMAP, where \frameworkname{} records EACS scores of 0.9932 and 0.9734, respectively. This performance sets a stark contrast to other models, such as MSCRED, which achieves commendable F1 scores, but  experiences a notable decrease in EACS due to prolonged training times (e.g., 774.99 seconds for MBA). In this light, \frameworkname{} distinguishes itself by adeptly optimizing both model complexity and training efficiency, showcasing its ability to balance high-performance metrics with operational practicality. The latter is particularly evident in datasets with challenging environments like SWaT and WADI, where \frameworkname{} not only sustains high F1 scores but also significantly reduces training times, achieving EACS scores of 0.9297 and 0.7883, respectively. While models like GDN and TranAD perform well in specific datasets, they do not consistently match \frameworkname{} in training efficiency across the board. This evaluation emphasizes the importance of considering training time and model size when deploying practical anomaly detection models.

\begin{table}[h!]
\centering
\footnotesize
\begin{adjustbox}{width=\textwidth, center, captionabove={Comparative analysis of the Efficiency-Accuracy-Complexity Score (EACS) across various anomaly detection models, highlighting the performance of \frameworkname{}. The EACS is calculated by normalizing and weighting F1 scores (0.4), training times (0.4), and parameter counts (0.2), demonstrating the balanced efficiency and accuracy of \frameworkname{} against benchmark models across multiple datasets}, label=Tab: EACS, nofloat=table}
\begin{tabular}{lcccc|cccc|cccc|cccc}
\toprule
\multicolumn{1}{c}{\textbf{Method}} & \multicolumn{4}{c}{\textbf{NAB}} & \multicolumn{4}{c}{\textbf{UCR}} & \multicolumn{4}{c}{\textbf{MBA}} & \multicolumn{4}{c}{\textbf{SMAP}} \\
 & F1 \% & Training Time (sec) & \# Params & EACS & F1 \% & Training Time (sec) & \# Params & EACS & F1 \% & Training Time (sec) & \# Params & EACS & F1 \% & Training Time (sec) & \# Params & EACS \\
\midrule
LSTM-NDT & 65.31 & 23.40 & 1710 & 0.8424 & 52.31 & 11.14 & 1765 & 0.7985 & 94.56 & 47.80 & 18641 & 0.9520 & 78.79 & 36.43 & 29146 & 0.8459 \\
DAGMM & 74.53 & 64.50 & 1266 & 0.8467 & 53.37 & 27.50 & 1214 & 0.7919 & 96.83 & 92.36 & 2448 & 0.9394 & 88.80 & 26.60 & 7266 & 0.9050 \\
OmniAnomaly & 74.42 & 88.70 & 13717 & 0.8250 & 83.46 & 38.21 & 13025 & 0.8863 & 92.25 & 197.66 & 25474 & 0.8649 & 87.28 & 66.77 & 16813 & 0.8230 \\
MSCRED & 75.02 & 503.60 & 1237377 & 0.3001 & 54.41 & 559.60 & 128655 & 0.2176 & 96.23 & 774.99 & 2441778 & 0.3849 & 86.64 & 25.70 & 8237452 & 0.6982 \\
MAD-GAN & 77.52 & 53.90 & 838 & 0.8671 & 85.38 & 31.50 & 822 & 0.9177 & 96.89 & 229.78 & 1877 & 0.8688 & 86.54 & 41.26 & 6718 & 0.8683 \\
USAD & 74.42 & 65.40 & 1359 & 0.8455 & 89.52 & 54.23 & 1359 & 0.9172 & 94.43 & 191.26 & 1609 & 0.8789 & 84.19 & 49.22 & 7395 & 0.8439 \\
CAE-M & 79.68 & 33.60 & 7229 & 0.8909 & 69.81 & 22.10 & 7365 & 0.8520 & 91.54 & 89.66 & 15411 & 0.9186 & 88.27 & 212.51 & 7229 & 0.5529 \\
GDN & 79.98 & 131.60 & 574 & 0.8153 & 68.94 & 64.70 & \textbf{566} & 0.8286 & 93.32 & 203.45 & 1106 & 0.8682 & 85.18 & 104.50 & \textbf{2974} & 0.7440 \\
TranAD & 93.64 & 3.37 & 615 & 0.9718 & 94.07 & \textbf{1.14} & 619 & 0.9745 & 97.80 & 5.11 & 1298 & 0.9885 & 89.15 & 5.62 & 62271 & 0.9445 \\
\frameworkname{} & \textbf{94.11} & \textbf{2.70} & \textbf{285} & \textbf{0.9742} & \textbf{99.10} & 2.24 & 1690 & \textbf{0.9922} & \textbf{98.61} & \textbf{2.27} & \textbf{593} & \textbf{0.9932} & \textbf{94.60} &\textbf{ 3.19} & 39045 & \textbf{0.9734} \\
\midrule
\multicolumn{1}{c}{\textbf{Method}} & \multicolumn{4}{c}{\textbf{MSL}} & \multicolumn{4}{c}{\textbf{SWaT}} & \multicolumn{4}{c}{\textbf{WADI}} & \multicolumn{4}{c}{\textbf{SMD}} \\
 & F1 \% & Training Time (sec) & \# Params & EACS & F1 & Training Time (sec) & \# Params & EACS & F1 \% & Training Time (sec) & \# Params & EACS & F1 \% & Training Time (sec) & \# Params & EACS \\
\midrule
LSTM-NDT & 77.21 & 37.80 & 61856 & 0.8887 & 66.70 & 41.50 & 72854 & 0.7957 & 2.71 & 422.67 & 133712 & 0.5753 & 90.42 & 460.70 & 42853 & 0.9100 \\
DAGMM & 84.82 & 25.66 & 18756 & 0.9258 & 81.28 & 29.66 & 16558 & 0.8748 & 14.12 & 246.32 & 39387 & 0.6498 & 94.91 & 337.90 & 10516 & 0.9420 \\
OmniAnomaly & 87.65 & 29.69 & 39753 & 0.9348 & 81.31 & 40.20 & 36541 & 0.8568 & 42.60 & 361.60 & 39753 & 0.7794 & 94.01 & 311.20 & 22941 & 0.9412 \\
MSCRED & 93.63 & 40.70 & 18476728 & 0.7535 & 80.72 & 236.80 & 14227264 & 0.3229 & 37.41 & 1884.25 & 75530 & 0.6716 & 84.14 & 349.77 & 14237356 & 0.6978 \\
MAD-GAN & 91.69 & 32.64 & 17446 & 0.9497 & 80.65 & 34.70 & 15958 & 0.8638 & 35.88 & 321.20 & 35764 & 0.7500 & 91.50 & 424.60 & 9903 & 0.9188 \\
USAD & 88.22 & 29.60 & \textbf{14859} & 0.9374 & 81.43 & 35.81 & \textbf{12502} & 0.8651 & 30.56 & 389.72 & \textbf{32859} & 0.7214 & 94.95 & 328.77 & 10609 & 0.9432 \\
CAE-M & 87.33 & 774.60 & 204687 & 0.5471 & 81.01 & 71.22 & 194525 & 0.8010 & 41.17 & 7724.92 & 388905 & 0.3503 & 93.67 & 3606.90 & 114219 & 0.5731 \\
GDN & \textbf{95.91} & 121.80 & 153541 & 0.9191 & 81.01 & 83.55 & 115442 & 0.7813 & 42.60 & 6047.12 & 287082 & 0.4559 & 83.42 & 1000.40 & \textbf{7255} & 0.8226 \\
TranAD & 94.94 & 6.88 & 272181 & 0.9733 & 81.53 & 2.10 & 31336 & 0.9221 & 49.51 & \textbf{177.60} & 1378173 & 0.6645 & 96.05 & 56.70 & 135110 & 0.9760 \\
\frameworkname{} & 94.82 & \textbf{5.44} & 236491 & \textbf{0.9739} & \textbf{83.14} & \textbf{1.60} & 13464 & \textbf{0.9297} & \textbf{84.35} & 227.50 & 1701515 & \textbf{0.7883} & \textbf{99.81} & \textbf{52.40} & 132050 & \textbf{0.9916}\\
\bottomrule
\end{tabular}
\end{adjustbox}
\end{table}

\begin{table}[h!]
\centering
\footnotesize
\begin{adjustbox}{width=\textwidth, center, captionabove={Hyperparameter Optimization Results: Detailed summary of the optimal training and architectural hyperparameters identified for each dataset, demonstrating the adaptive precision of \frameworkname{} in time series anomaly detection.}, label=tab:Best model params, nofloat=table}
\begin{tabular}{|l|l|l|l|l|l|l|l|l|}
\hline
\textbf{Arch and Training Params} & \textbf{SMAP} & \textbf{UCR} & \textbf{MBA} & \textbf{SWaT} & \textbf{MSL} & \textbf{SMD} & \textbf{NAB} & \textbf{WADI} \\ \hline
Learning Rate & 0.0002128 & 0.0019997 & 0.003441 & 0.0000403 & 0.006192 & 0.0002273 & 0.006924 & 0.0015406 \\ \hline
Dropout Rate & 0.2353 & 0.4474 & 0.1795 & 0.3836 & 0.3730 & 0.1554 & 0.4560 & 0.2886 \\ \hline
Dim Feedforward & 101 & 124 & 41 & 42 & 121 & 86 & 10 & 24 \\ \hline
Batch Size & 32 & 48 & 16 & 32 & 32 & 48 & 96 & 128 \\ \hline
Encoder Layers & 2 & 1 & 3 & 2 & 2 & 3 & 1 & 2 \\ \hline
Decoder Layers & 1 & 2 & 1 & 3 & 3 & 1 & 2 & 1 \\ \hline
Activation Func. & sigmoid & leaky\_relu & tanh & tanh & leaky\_relu & sigmoid & relu & leaky\_relu \\ \hline
Time Warping & False & True & True & True & True & False & False & True \\ \hline
Time Masking & True & True & False & True & True & False & True & False \\ \hline
Gaussian Noise & 0.000151 & 0.027956 & 0.007925 & 0.058019 & 0.000428 & 0.000628 & 0.000119 & 0.001042 \\ \hline
Linear Embedding & False & True & False & False & True & True & True & True \\ \hline
Phase Type & 2phase & iterative & iterative & 2phase & 2phase & 2phase & iterative & 2Phase \\ \hline
Self Conditioning & False & True & False & True & True & False & True & False \\ \hline
Layer Norm & False & False & False & True & True & True & False & False \\ \hline
Pos. Enc. Type & sinusoidal & fourier & fourier & sinusoidal & fourier & sinusoidal & sinusoidal & sinusoidal \\ \hline
FFN Layers & 1 & 1 & 1 & 3 & 2 & 3 & 1 & 1 \\ \hline
Attn Heads & 25 & 1 & 2 & 51 & 55 & 38 & 1 & 127 \\ \hline
Window Size & 10 & 20 & 14 & 22 & 26 & 12 & 18 & 26 \\ \hline
\end{tabular}
\end{adjustbox}
\end{table}

\subsection{Optimal Model Configurations Identified by \frameworkname{}}

Table \ref{tab:Best model params} showcases the culmination of \frameworkname{}'s NSGA-II optimization process, highlighting the training and architectural hyperparameters of models that achieved the best balance between high F1 scores and low parameter counts. These configurations represent models uniquely suited to each dataset's anomaly detection needs.

A key observation is the diversity in architectural parameters across datasets, reflecting \frameworkname{}'s adeptness in tailoring models to specific data characteristics. For example, in the SMAP dataset, the model employs a relatively simple architecture with fewer encoder and decoder layers, which is effective in handling the environmental time series data of this dataset. In contrast, the WADI dataset, known for its complex sensor network, necessitates a more intricate model structure, evident in its higher number of attention heads and the use of a two-phase reconstruction and refinement strategy.

Similarly, the variation in window sizes, ranging from 12 data points for SMD to 30 data points for WADI, underscores the model's flexibility in adapting to the temporal scale of different datasets. Larger window sizes in datasets like WADI allow the model to capture longer-term dependencies and subtle anomalies in extensive time series data, a crucial requirement for sophisticated water-related infrastructures.

The adaptability of \frameworkname{} is further demonstrated in its choice of positional encoding and layer normalization techniques, which vary significantly among datasets. For instance, the sinusoidal positional encoding in SMAP and NAB caters to their unique temporal patterns, while the Fourier positional encoding in UCR and WADI aligns with the datasets' complex spectral characteristics.

These results from \frameworkname{}'s NAS process validate the framework's ability to not only fine-tune hyperparameters for operational efficiency, but also to intricately adapt its architectural design to the nuanced demands of various time series anomaly detection scenarios, ensuring optimal detection performance across a broad spectrum of datasets.

\begin{figure}[!ht]
\centering
\includegraphics[width=1.0\textwidth]{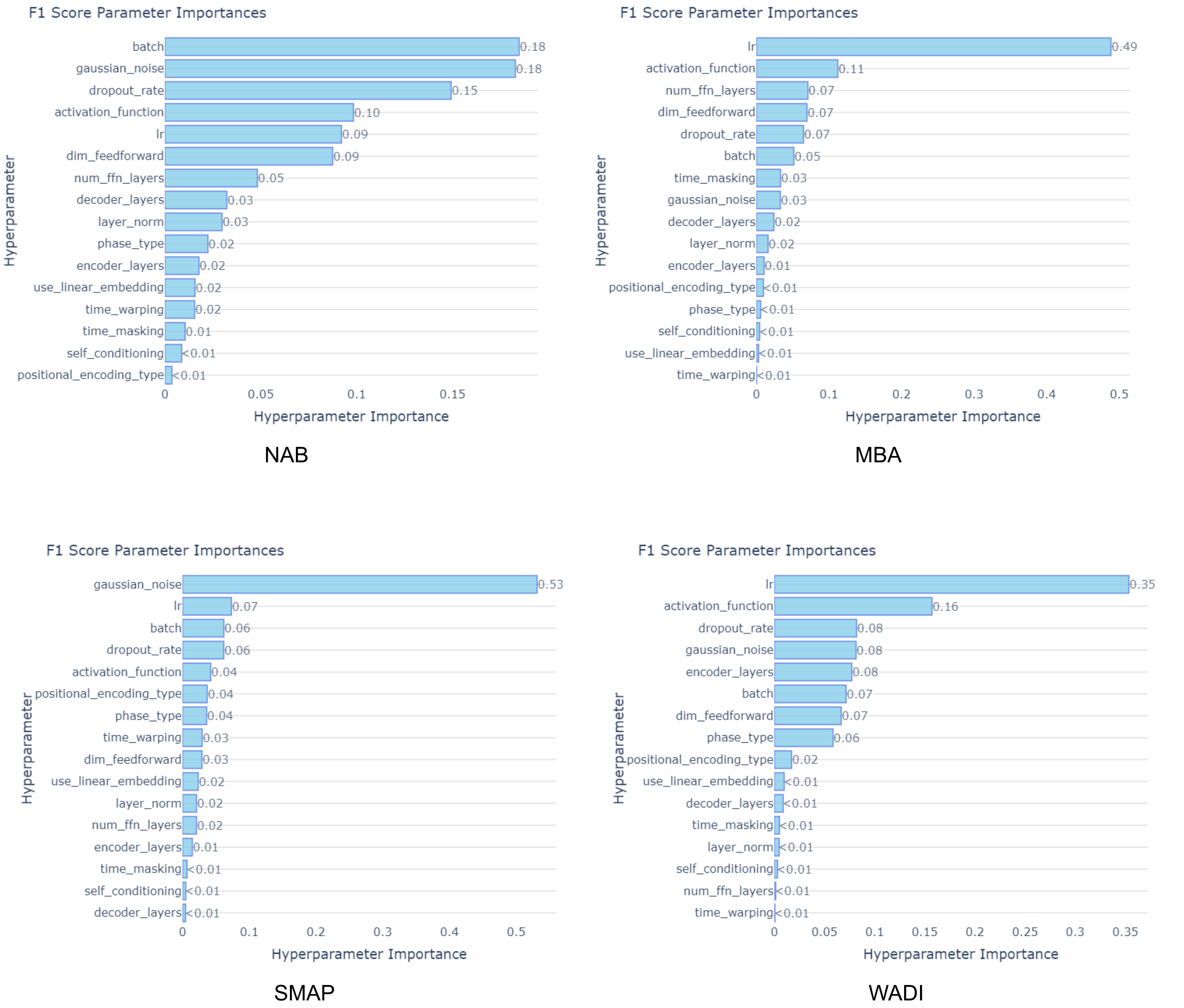}\\
\caption{Analysis of hyperparameter importance for F1 score optimization in \frameworkname{} across four datasets (NAB, MBA, SMAP, and WADI). The plots illustrate the relative impact of different hyperparameters on the F1 score, guiding the model's fine-tuning process for effective anomaly detection in diverse time series datasets}
\label{fig:hyeperparam-Imp}
\end{figure}

\subsection{Architectural and Training Hyperparameter Importance for F1 Score Optimization}

The optimization of F1 scores in the \frameworkname{} model shown in Figure~\ref{fig:hyeperparam-Imp} is critically analyzed through hyperparameter importance plots for four datasets: NAB, MBA, SMAP, and WADI. These plots offer quantitative insights into the relative impact of various hyperparameters on the F1 score, aiding the fine-tuning process in our NAS-driven anomaly detection model.

For the NAB dataset, \texttt{gaussian\_noise} and \texttt{activation\_function} are the most influential hyperparameters, with importances of 0.18 and 0.15, respectively. The incorporation of \texttt{gaussian\_noise} enhances the model's generalization capabilities, essential for robustness, while the choice of \texttt{activation\_function} is key for capturing the data's nonlinear relationships.

The MBA dataset's analysis underscores the importance of \texttt{activation\_function} as 0.11, \texttt{num\_feedforward\_layers} as 0.07, and \texttt{dropout\_rate} as 0.07, indicating the need for a complex model architecture to effectively learn from multivariate ECG time series.

In the SMAP dataset, \texttt{gaussian\_noise} with importance 0.53  highlights the model's ability to handle noisy environmental data effectively. Additionally, \texttt{batch\_size} with importance 0.07 influences the model's performance, reflecting the impact of batch processing on training.

The WADI dataset, with its extensive sensor network, prioritizes \texttt{activation\_function} with importance 0.16 and \texttt{dropout\_rate} with importance 0.08 to maintain model robustness and prevent overfitting in high-dimensional spaces.

The recurrence of \texttt{activation\_function} and \texttt{dropout\_rate} across datasets emphasizes their role in non-linear data transformation and regularization. This pattern reflects the impact of data quality and structure on the model's learning efficacy.

This hyperparameter analysis illuminates the tailored performance of \frameworkname{} across varied datasets, with each dataset's specifics dictating the importance of particular hyperparameters. This not only affirms the model's adaptability but also its capacity for ongoing enhancement, ensuring it remains adept at confronting the evolving complexities of time series data.

\begin{figure}[ht]
\centering
\includegraphics[width=0.8\textwidth]{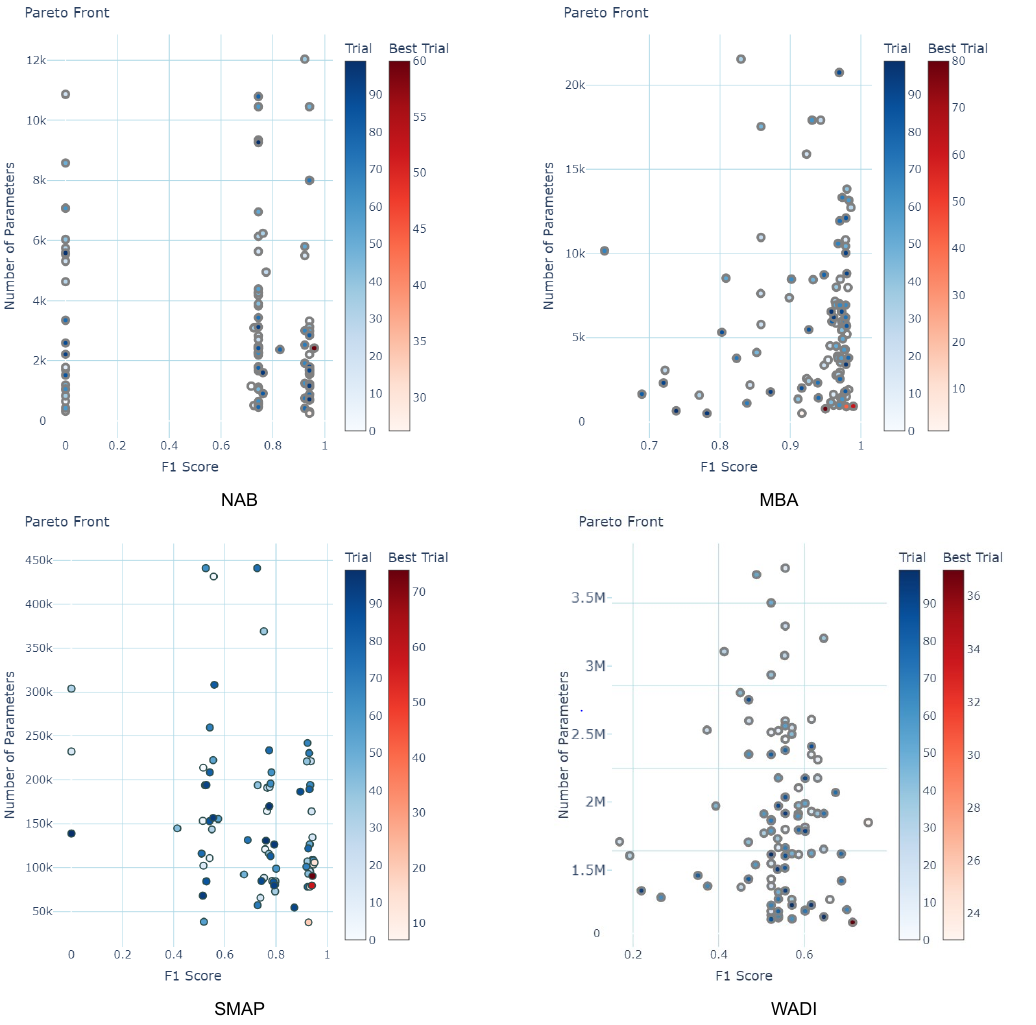}\\
\caption{Pareto front plots illustrating the trade-off between F1 score and number of parameters for NAB, MBA, SMAP, and WADI datasets in the \frameworkname{} optimization process.}
\label{fig:Pareto-front}
\end{figure}

\subsection{Pareto Front Analysis for Model Optimization}

Building upon the insights from the hyperparameter importance analysis, the Pareto front optimization results for the \frameworkname{} model are depicted in Figure~\ref{fig:Pareto-front}. The Pareto front plots for the NAB, MBA, SMAP, and WADI datasets underscore the delicate balance between model complexity, as measured by the number of parameters, and the model's effectiveness, as quantified by the F1 score.

For the NAB dataset, the Pareto front indicates a dense congregation of model configurations. A significant number of these configurations exhibit a high parameter count without a corresponding increase in the F1 score, suggesting a potential plateau in performance gains relative to complexity. This observation prompts a critical evaluation of model parsimony, encouraging the selection of simpler models that maintain performance while mitigating the risk of over-fitting.

In the MBA dataset, the dispersion of trials across the Pareto front reflects a comprehensive exploration of the architectural space. Interestingly, several models achieve commendable F1 scores without a proportionate surge in parameters, highlighting efficient architectural choices that capture the essential dynamics of ECG time series data without unnecessary complexity.

The SMAP dataset presents an outlier model with a substantial number of parameters yet achieving a high F1 score. This result may point to an overfitting scenario where the model's complexity does not translate into generalized performance. Such an insight is invaluable for guiding the model selection process towards architectures that balance accuracy with generalization.

The Pareto front for the WADI dataset, characterized by its complex sensor network, illustrates the necessity of sophisticated models to enhance the F1 score. The trend of increasing model complexity to achieve incremental improvements in performance is evident, underlining the inherent challenges of anomaly detection in high-dimensional industrial control systems.

These Pareto fronts provide a visual and quantitative tool for identifying optimal model configurations that achieve a balance between performance and complexity. They serve as a decision-making aid for selecting models that  align with the practical demands of anomaly detection in diverse environments.

The findings from the Pareto analysis are integral to the ongoing development of the \frameworkname{} model within the broader scope of our research. They contribute to understanding how different model architectures perform across various datasets, and offer a pathway to enhancing the model's adaptability and ensuring its continued efficacy in the face of evolving data challenges.

\section{Discussion: Addressing Challenges and Exploring Future Directions in \frameworkname{}}

\frameworkname{} represents a significant advancement in anomaly detection within time series data, primarily through its innovative integration of advanced adversarial learning paradigms, NSGA-II optimization, and transformer architecture optimization. This confluence of technologies marks a substantial leap forward, particularly in the detection of subtle and complex anomalies that have typically eluded traditional methods.

A pivotal challenge addressed by \frameworkname{} is the balance between detection accuracy and computational efficiency. The NSGA-II algorithm plays a critical role in optimizing model performance, without incurring excessive computational demands. This aspect is particularly crucial in real-world applications where resources are finite and efficiency is paramount.

The adaptability of \frameworkname{} is further exemplified in its dynamic adjustment capabilities, allowing it to effectively respond to the unique characteristics of different datasets. This adaptability is essential in the domain of anomaly detection, where dataset variability can present diverse challenges. Additionally, the iterative self-adversarial approach employed by \frameworkname{} significantly enhances detection accuracy, showcasing the model's sophisticated capabilities in identifying anomalies.

However, challenges remain, particularly in the realm of ensuring model generalization across a diverse array of datasets. The Pareto front analysis within \frameworkname{} has highlighted a delicate balance between model complexity and effectiveness. Some configurations risk overfitting, which is a pertinent issue for future research endeavors. Improving generalization capabilities, without compromising detection accuracy, remains a key area for further investigation.

Looking ahead, several promising avenues for enhancement and innovation present themselves. Enhanced real-time data processing capabilities, particularly for applications in environmental monitoring and industrial control systems, represent a significant area for advancement. Techniques from data assimilation and online learning could be effectively integrated into \frameworkname{} to address these challenges. Additionally, the development and implementation of dynamic thresholding strategies, such as the Moving Average Thresholding (MAT) and the incorporation of rolling statistics, offer exciting prospects. 

Furthermore, the exploration of hybrid systems that synergize machine learning with simulation approaches, along with advancements in neuro-symbolic systems, could substantially enhance the model’s adaptability and effectiveness across various scenarios. Finally, a human-centric approach to machine learning, integrating human feedback in a more intuitive and formalized manner, remains a significant challenge. \frameworkname{} stands to benefit greatly from such integration, enhancing not only the interpretability but also the overall usability of the model in real-world applications.

As the landscape of time series anomaly detection continues to evolve, so too will the strategies and methodologies embodied in \frameworkname{}, ensuring its continued relevance and efficacy in this dynamic field.

\section{Conclusion}




\frameworkname{} signifies a notable contribution in the realm of time series anomaly detection by merging transformer architecture with neural architecture search and NSGA-II optimization, achieving superior performance across diverse univariate and multivariate datasets. Its robustness in accurately detecting anomalies highlights the effectiveness of this integration, effectively addressing the dual challenge of maintaining accuracy while ensuring computational efficiency—key for practical applications. This framework is further distinguished by its incorporation of advanced adversarial learning paradigms, enabling the detection of nuanced anomalies and marking a significant step forward from traditional methods. As the landscape of data analysis evolves, \frameworkname{} not only sets new performance benchmarks in anomaly detection but also opens up exciting avenues for future research, particularly in enhancing real-time processing and integrating human-centric approaches to machine learning. The principles and approaches embodied in \frameworkname{} are paving the way for innovative applications across various sectors, shaping the future of machine learning with its groundbreaking methodologies.

\section*{Acknowledgments}

This material is based upon work supported by the National Science Foundation under Grant No. EAR 2012123. Any opinions, findings, and conclusions or recommendations expressed in this material are those of the author(s) and do not necessarily reflect the views of the National Science Foundation. Any use of trade, firm, or product names is for descriptive purposes only and does not imply endorsement by the U.S. Government.
The work was also supported by the University of Vermont College of Engineering and Mathematical Sciences through the REU program.
We would like to extend our gratitude to the providers of the data sets used in this work, whose contributions were invaluable to our research.





\end{document}